# Hybrid Convolutional Neural Networks with Reliability Guarantee

Hans Dermot Doran
*Institute of Embedded Systems*
*Zurich University of Applied Sciences*
Winterthur, Switzerland
donn@zhaw.ch

Suzana Veljanovska
*Institute of Embedded Systems*
*Zurich University of Applied Sciences*
Winterthur, Switzerland
veln@zhaw.ch

*Abstract*— **Making AI safe and dependable requires the generation of dependable models and dependable execution of those models. We propose redundant execution as a well-known technique that can be used to ensure reliable execution of the AI model. This generic technique will extend the application scope of AI-accelerators that do not feature well-documented safety or dependability properties. Typical redundancy techniques incur at least double or triple the computational expense of the original. We adopt a co-design approach, integrating reliable model execution with non-reliable execution, focusing that additional computational expense only where it is strictly necessary. We describe the design, implementation and some preliminary results of a hybrid CNN.**

## I. INTRODUCTION

Our aim is to investigate the construction and characterisation of neural networks, in this case convolutional neural networks, which perform execution of some convolution layers reliably. Our body of work is therefore a contribution to the field of dependable machine learning. We target our output at (embedded) edge-AI devices towards applications where functional safety or safety of the intended function is required.

We structure the rest of the paper accordingly. We finish this section with a review of related work and some statements to our novelties. In Section II we discuss some reliability techniques. In Section III we introduce the concept of the hybrid neural network. In Section IV we discuss the implementation before concluding in Section V with some remarks and suggestions for further work.

### A. Related Work

The fact that AI is not inherently safe has not escaped attention [1]. There is however a wide interpretation of the meaning and scope of the term "safety". The effect of lax security environments [2] inspires [3] to define open challenges of robustness, monitoring, alignment and systemic safety requiring immediate attention. Many authors identify the well-known adversarial perturbation (robustness) problem [4] as a major obstacle to safety in AI, particularly in important applications such as autonomous driving [5] and medical applications. Safety and trust are often brought together in this context [6] as is explainable AI [7], [8], [9]. We align ourselves with relevant standardisation work including IEC 61508 [10] and ISO 26262 [11] and we use the terms dependability, availability, and reliability in the definition context of the IEC 60500-192:2015 [12]. The literature recognising these standards, whilst sparser, also focuses on the domains of autonomous vehicles and medical systems [13], [14]. Literature targeting compliance with specific standards are [15], [16], transport [17], [18] and medical [19].

Current functional safety thinking recognises failure probabilities rather than certainty of operation and this is reflected in AI/safety related research, for instance [20]. For practical robotic applications using reinforcement learning, supervision [21], [22], [23], [24] is a typical approach. AI-hybridization [25] has also been suggested, in this case the authors suggest combining data- and knowledge-driven AI as a supervisor for automated driving functions. Caging of some sort, that is where an output is bounded or undergoes some plausibility check, is another favoured approach. This can occur on a modelling layer by the use of surrogate functions [26], on classification [27], or after individual operations [28]. The latter masks failures whereas the two former detect failures. We reliably execute some CNN layers. The outputs bifurcate, serving as inputs to a (reliable) feature determination via a bounded surrogate function as well as subsequent layers of the CNN. The results of the feature determination are used to qualify the identification of certain classes by the CNN.

In embedded applications, edge-AI is typically understood as a low-power and low-computational power domain. AI accelerators, such as the Nvidia Jetson series [29] feature ~128 cores per chip, far removed from the high performance computing (HPC) world. Nevertheless the HPC world also suffers from similar reliability issues as edge-AI so the solutions come from the same solution set [30]. Specifically, the cost of computation on the edge and in HPC warrants the use of checkpointing and rollback techniques. We shall discuss these techniques in subsequent sections but we note here that we apply a form of checkpointing and rollback.

### B. Novelties

Our novelty is to present the concept and some work on a hybrid (convolutional) neural network to facilitate reliable neural network execution towards safe and dependable AI. In this context we apply checkpointing and rollback on an operation-execution basis, to conserve the value of previous

We thank SERI, Innosuisse and the ZHAW for the financial support of this research under the umbrella of the KDT project REBECCA, grant agreement n° 101097224.

executions. Only persistent failures are explicitly reported. Our proposal should eventually execute on field programmable gate array (FPGA) technology. To preserve generality we take a generic, software based approach rather than implementing for specific accelerator architectures.

## II. TECHNIQUES FOR RELIABLE EXECUTION

Reliability, the ability of a circuit to function without failure, is a key factor in functional safety, especially safety functions, as these must exhibit a low probability of failure on demand. The failure of a number of calculations in a CNN due to single event upsets acting on the processing element or data corruption of the weights and input data may critically alter the result [31]. The target for reliability engineering is to ensure that any calculations are carried out correctly, and typically within time-limits. If errors do occur, these are to be caught to prevent error propagation throughout the system.

### A. Lockstep Processing

A well-known example of a reliable circuit is a tightly-coupled lockstep processer, consisting of two (or more) microprocessors operating in lock-stepped mode, that is, executing the same machine code within two or less clock-cycles of each other. The activities of the processors that are visible on the system bus are compared and should the two not be the same, for example one attempts a code fetch from a different address, an error is flagged. Lockstep processors are often used in embedded safety-critical systems such as automatic braking systems (ABS) in automotive engineering [32] or fieldbus-connected emergency-off buttons in factory automation applications [33]. Errors-causing failures are typically exampled to be radiation-caused single event upsets (SEU) which are random, the assumption being that such an error is unrecoverable and will not be present once the system has re-booted. In the case of an ABS system, control is effectively returned to the driver. When an error is detected in such a unit, a system-reset is generally carried out [34].

In many applications a system-reset is not an appropriate response to a single error. For instance, in a parallel set of calculations carried out on a compute unit[1] such as a graphics processing unit (GPU) or a neural processing unit (NPU), it is self-evident that the failure of one of 128 processing elements[2] (PE), causing a total safety-relevant system shutdown cannot be considered desirable.

### B. Processing with Redundancy

This concern has long been considered in the domain of software reliability engineering and multicore architectures where checkpointing and rollbacks in the case of errors are common [37]. Detection whether an error has occurred usually involves redundant execution. This can be achieved on a spatial basis using for instance two otherwise independent compute units. Two independent compute units implies that execution proceeds asynchronously and synchronisation delays for result comparison apply. Redundancy can also be achieved on a temporal basis, the use of one compute unit with the same software executed twice in series which implies at least twice the execution latency to reach a given deadline vis a vis a single execution. The advantage of checkpointing is that an explicit barrier to error propagation is erected and yet a considerable number of instructions can be executed between checkpoints. The disadvantage is that in a repetitive error case, there are few mechanisms available to halt rollback and re-execution. In the case of temporal redundancy and given a permanent error, the platform becomes unusable. In the case of spatial redundancy and (also) given an error, the platform has the potential to operate in a reduced mode allowing the implementation of graceful degradation strategies.

### C. Lessons from High Performance Computing

A parallel processing compute unit such as a GPU or a NPU features spatially independent and schedulable processing elements and so is predestined, in the context of reliability engineering, to execute both spatially and temporally redundant operations. Both of the largest GPU vendors active in the embedded market provide features supporting reliability towards functional safety in their offerings [38], [39] but these features are not always fully public. There is a need for vendor independent examination of reliability constructs for GPU/NPU compute units. Reliability analysis in high performance computing (HPC) clusters is well understood, GPUs with their ability to coalesce multiple execution streams on one hardware, less so. As with most compute device architectures reliability analysis focuses either on memory or the processing elements. GPU manufacturers have begun implementing error correcting codes in RAM storage and data paths, both DRAM and cache-SRAM, and it can be shown that more vendor attention is being paid to hardening on-chip SRAM structures [40]. Understanding GPU processing element hardness (to soft errors) is still very much a work in progress [41]. Diverse execution of redundant GPU kernels has been examined [42] but programming GPU kernels is a very architecture-dependent activity and it cannot be presumed that the GPU architecture will become the preferred AI accelerator architectures in edge-AI applications. One of the novelties we offer in this body of work is an architecture-independent framework for redundant execution in AI applications from which architecture-dependent derivatives can be built.

### D. Checkpoints and Rollbacks

Checkpointing as described above is an approach that has been suggested in a variety of embodiments. Since the potential numeric output space of a CNN classifier is very large, even if crudely hidden by function with codomains such as the activation and softmax functions, placing bounds on the output space appears sensible. The "caging" variant, proposed by [27] checks for output feasibility against a permissible output space. Another caging variant checks the outputs of operations and if they are larger or smaller than some preset and operation specific saturation limit, the output saturates to that value [28]. Whilst this approach preserves computing power vis a vis redundant execution, the required memory bandwidth is substantially increased as not only must the weights and data be loaded from memory but also the upper and lower saturation limits.

---

[1] Here we use OpenCL [35] terminology to denote a compute device consisting of one or more compute units including one or more processing elements.

[2] in the case of the Jetson Nano [36], a GPU explicitly marketed for edge-AI applications

### E. Proposal

Execution of an entire CNN in duplicate with intermittent checkpointing represents, in an embedded application, a significant computational expense that may be optimisable. A rollback to a checkpoint and re-execution represents a significant delay to output of results. Once there are hard or soft deadlines to be met, the rollback-distance becomes a significant consideration. The rollback-distance is understood to exist in a synergistic relationship between hardware and software [43]. In a convolution layer which consists largely of multiplications and additions, the rollback-distance can be reduced to one operation. That is, for instance, a redundantly executed multiplication with result comparison (checkpoint) and a re-multiplication (rollback) should the first have failed. The concept of a reduced number of redundant execution layers has been previously examined and found wanting, [44] concluding that it is not possible to guarantee reliable execution and any partial reliability that may be achievable is highly dependent on the specific network architecture.

In both the cases of [27], [28], and that which we are presenting here, the teaching of an algorithm must be subjected to an additional workflow step to determine the output bounding set. This in turn requires an understanding of the network algorithm, trained model and especially the nature of the data with which the model has been trained. It has long been recognised that training data quality is reflected in the quality, however defined, and robustness of the machine trained models [45]. Since then work in this domain has been sparse, although there has been more recent activity in the context of machine learning workflows [46] and explainable, responsible and trustworthy AI [47], [48]. In our view a hybrid (reliable/non-reliable) network offers advantages in that any data use to train or otherwise modify the model weights for reliability purposes should benefit the other segments of the model. This is not the case in the caging solutions already mentioned. We therefore offer a substantial and further explorable novelty in this regard.

### III. Hybrid (Convolutional) Neural Network Design

A (convolutional) neural network may be understood as a mapping function, each layer mapping input data to a set of dimensions. Adding reliability to this mapping function is possible but computationally expensive. Our insight is that not all classifications may be relevant for reliability purposes and hence not all layers or portions of layers need be executed reliably. By viewing this as a data flow problem rather than a control flow problem it is possible to imagine a partitioning of the neural network into reliable and non-reliable executions. We explain this concept further in the next subsection.

#### A. Partitioning of a Reliable CNN

Our experimental use-case is the recognition of a "Stop" traffic sign. We pick this sign because it contains redundant information including the shape. The octagonal shape of the sign, recognisable from the front and the back, is qualified as relevant for the approaching driver by colour. Recognising the shape and colour therefore is paramount to achieving a true positive and avoiding either a false negative or false positive. This in turn implies that the shape must be captured in its entirety, together with any relevant context. This requirement effectively means that any shape recognised by a CNN is not a "Stop" sign unless the shape has been confirmed as octagonal. This insight informs the dataflow of definitive recognition of a "Stop" sign in the context of a CNN. The simplest form (Figure 1) is to maintain a shape-recognition functional block in parallel with a CNN for a general classification. The output signal of the qualifier will qualify a single relevant output of the CNN and indicate this in the "Reliable Result" block. This method is only really feasible in applications where a significant percentage of the classifier output must not be dependable and there exists a method to reliably determine specific qualifying aspects of the input signal set. For instance, if the classifier detects a "Stop" sign then that determination can be qualified by the output of a reliably executed shape recognition block. Classifications that are not considered safety critical (e.g., a parking prohibition) can be used without any qualification. Using a qualifying signal is a good way of assessing a signal of non-dependable provenance for formal correctness (adherence to signal timeliness, voltage/current limits, time-characteristics (baud-rates) etc.) thus raising the apparent probability that the delivered signal is feasible.

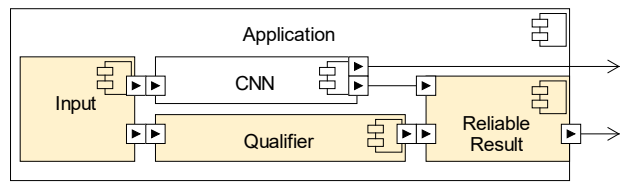

*Figure 1: Qualification of a CNN classification using a reliably executed qualifying block.*

Machine learning seeks to extract generic features from a general dataset and dependability seeks to determine definitive statements based on well understood data sets. It would appear intuitive to allow the machine learning to benefit from any data sets the dependability-specific processing blocks utilise. Integrating this intuition into a CNN structure facilitates this body of though shown in Figure 2. This figure shows that some subset of the CNN, the dependable CNN (DCNN) is to be executed reliably. The data path will eventually bifurcate as the CNNs task is to generalise and the reliable executions task is to become deterministically specific. How the CNN may benefit from the data used to construct the qualifier will be discussed in the next section. This architecture we refer to as a hybrid CNN.

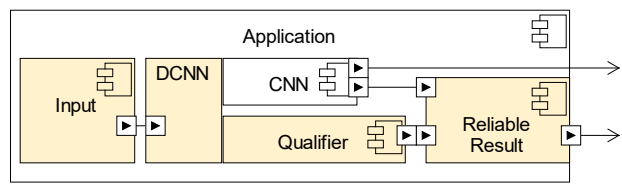

*Figure 2: A hybrid CNN featuring reliable (DCNN) and non-reliable (CNN) execution.*

## B. Data Set Integration

In this body of work, we are focused on presenting the mechanics of integration of dependability features into a CNN. We are not examining the semantics of integrating definitively ascertained and inferred statements or positions (e.g., dependable data sets) into CNN training data. We therefore restrict ourselves to two, as it turns out, simple experiments to demonstrate our position. In our example case as explained, we have chosen a "Stop" sign recognition to experiment on. We determine the shape in the "Qualifier" block (Figure 2) by using a surrogate function [26] – whose upper and lower bounds can be determined a priori. This produces deterministic results that are fully explainable, for instance during a safety certification process. We use Symbolic Approximation (SAX) [49], which effectively reduces time-series data to a string which can be cheaply compared to other strings. The time-series of a shape is generated by the distance from the centroid of the shape to the edge of the shape. The time series of a slightly angled "Stop" sign (octagon) from the German traffic sign recognition benchmark (GTSRB) [50] is illustrated in Figure 3 below.

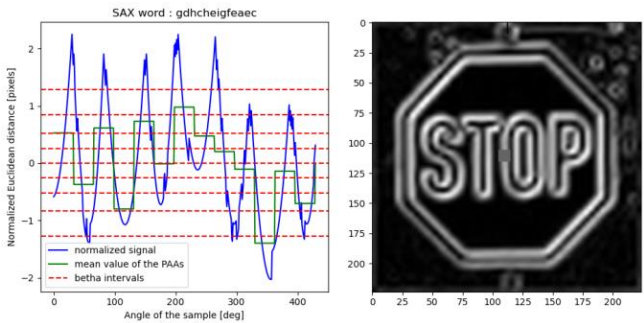

*Figure 3: The time-series generated from a real-world (GTSRB), slightly angled stop sign (top). The eight corners can be clearly identified. The SAX word is visible above the time-series plot.*

We wish to demonstrate the concept of a hybrid CNN using well-understood CNN algorithms, shape determination requires an appreciable image size with a clearly definable edge. We choose AlexNet [51] as this requires a barely acceptable for deterministic edge recognition, 227*227*3 input image. The first convolution layer of the AlexNet reduces the input using 96 11*11*3 filters. On an AlexNet trained with the GTSRB. As the SAX process requires an edge, we naively replace the first of the filters with a Sobel-x, Sobel-y, Sobel-x filter. We compare both the confusion matrices of the original and replaced filters and the accuracy and note no substantial difference in classification accuracy. Replacing all the 96 filters one at a time with the Sobel filters results in the plot of class confidence values shown below in Figure 4. The red dotted line in the plot indicates the accuracy of the original model. It is clearly visible that the accuracy varies substantially depending on which filter has been replaced.

Although this act does not satisfy our desire to allow the learning process to benefit from dependability-mandated data and is, as explained later, non-optimal from a CNN development point of view, it suffices as a baseline measurement.

We then begin pre-initializing one of the three-dimensional AlexNet filters to Sobel filters and train the network keeping this initalisation constant. In theory the training tool, TensorFlow in this case, offers the ability to freeze a filter during training. In practice, after every epoch or batch, the filter values are minimally changed, apparently to reflect the numeric balance of values presented to the pooling layer. It can be shown the (learnt) filter undergoes subtle changes in the intensity, statistical and spatial frequency domains. The accuracy of the model is not affected whether the kernels are replaced after training is completed or set before training has begun and re-set after every epoch or batch. It is thus not entirely clear whether the model can benefit from this pre-initalisation, in other words whether the training has benefited from the pre-initalisation, but in this case clearly exhibits no negative effects. Further investigations, out of scope of this body of work, will be appropriate on this point. For our purposes we have shown integration of a dependability-related feature in a workflow that is well understood in terms of CNN training.

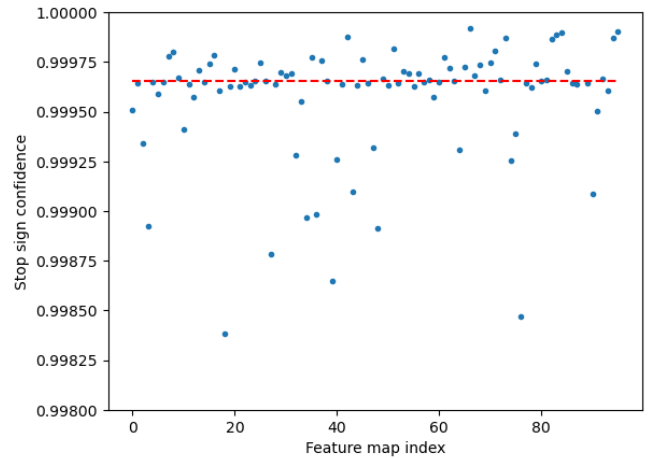

*Figure 4: Confidence values for the "Stop" sign class after replacement of each one of the learnt, first convolution layer AlexNet filters with a Sobel filter.*

## C. Closing Remarks

Having demonstrated fundamental feasibility of integrating dependability and non-dependable data paths in a CNN, thus creating a hybrid CNN, we progress onto implementation details. We postulate the determination of one (three dimensional) filter in the first convolutional layer. Further work will be required to investigate the manipulation of the subsequent layers in the CNN. Once the method of this determination is understood, the integration represents prima facia an optimization problem. Optimization with respect to balancing the generation and computational complexity of an application-specific qualification block with the general computational complexity of the CNN. Both, that is deterministic feature generation and integration optimization are clearly out of scope of this body of work and will be left to future investigations.

## IV. HYBRID NETWORK IMPLEMENTATION

As explained in the previous sections our intention is to create a hybrid CNN network, based on the AlexNet due to the larger image sizes that are required for shape recognition that are supported by AlexNet.

This algorithm has been designed to a number of expectations and constraints. Our final targets are FPGA devices. The FPGA workflow includes design of the hardware components in VHDL and simulation of these in GHDL [52] to which we bind cocotb [53], an open verification methodology (OVM) supporting verification tool which is geared towards Python verification code. We therefore can use the relatively high-level environment of Python to refine various versions of the algorithm. More importantly it allows us to attach an execution framework like TensorFlow to specially designed FPGA-instantiable circuits for verification and characterization purposes. There are a substantial number of degrees of freedom when implementing arithmetic operations in an FPGA, including use of integrated digital signal processing [54] elements, the use of high-level synthesis, the use of hand crafted VHDL code and the various architectures, reliable or not, for creating arithmetic operators. An exhaustive evaluation of these possibilities is out of scope of this body of work, so we suffice with a small Python implementation. We do however expect that the basic operators return a value, this value can be determined by a single execution of an operation or, in the case of triple modular redundancy, agreed upon by execution of the algorithm three times and voting on the result. The basic operators should also return a qualifier indicating whether the operation was carried out correctly or not. An example of this approach on a hardware level can be consulted in [55].

The algorithm depicted below (Algorithm 3) calculates one convolution operation. It assumes that every operation fails unless explicitly asserted otherwise. The exit conditions are failure or success, in this version we do not return diagnostic information other than maintain an error counter as a global variable. The algorithm executes an overloaded multiplication and an overloaded addition (accumulate), the two core operations of a convolution. If an error occurs during the execution of an operation, then, following the leaky bucket pattern [56], an error counter is incremented by a value (*factor* line 12) and checked against a ceiling,(defined line 2, checked line 12). For every correct operation this error counter is decremented by one, floor zero (line18-19). In this way a stream of correctly executed operations will cancel one, but not two successive errors. To increase availability, should one incorrect operation occur then that operation shall be repeated.

The overloading allows us to attach multiple methods to a basic operation. In Algorithm 1 we show a non-redundantly executed multiplication operation. This operation simply returns a product and a predefined qualifier, set to *True*. We use operations like this to determine baseline performance characteristics.

The next algorithm, Algorithm 2 shows a redundantly executed multiplication. Here the qualifier is set to *True* should the two products be the same. Other variants of reliable calculations can also be accommodated, possibly with slight variants to Algorithm 3. Note that in both cases the best-case execution and worst-case execution time are, given constant-time adders and multipliers, determinable and, in hardware, constant.

```
Single Multiplication Operation
1:  start procedure
2:      result ← k · i              → compute one calculation
3:      qualifier = True             → set the flag to True
4:      return qualifier, result     → return the flag and the result from the multiplication
5:  end procedure
```
*Algorithm 1: Example of non-reliably executed multiplication.*

```
Redundant Multiplication Operation
1:  start procedure
2:      result₁ ← k · i              → compute one product
3:      result₂ ← k · i              → repeat the calculation
4:      qualifier ← False            → set qualifier to false
5:      sub₁ ← result₁ - result₂     → compare the multiplications
6:      sub₂ ← result₁ - result₂
7:      if sub₁ = sub₂ and sub₁ = 0 then  → if the subtractions are 0, set the qualifier as True
8:          qualifier = True
9:      end if
10:     return qualifier, result₁    → return the flag and the result from the multiplication
11: end procedure
```
*Algorithm 2: Example of a reliably executed multiplication.*

```
Algorithm 1 Reliable Convolution Kernel
1:  error ← 0                        → initialization of global variables
2:  factor ← 8, n ← factor*2
3:  failure ← True
4:  start procedure
5:      failure ← True               → initialization of local variables
6:      sum ← 0, result ← 0, temp ← 0
7:      for i in kernel height do
8:          for j in kernel width do
9:              qualifier ← False
10:             qualifier, product ← k_{i,j} * i_{i,j}   → execute multiplication
11:             if qualifier = False then    → on error
12:                 if error ← error + factor > n then   → exit if error ceiling reached
13:                     result ← 0, break
14:                 qualifier, product ← k_{i,j} * i_{i,j}   → re-execute multiplication
15:                 if qualifier = False then
16:                     if error ← error + factor > n then
17:                         result ← 0, break
18:             if error > 0 then    → decrement error
19:                 error ← error – 1
20:             qualifier ← False
21:             qualifier, temp ← sum + product   → execute addition
22:             if qualifier = False then    → on error
23:                 if error ← error + factor > n then   → exit if error ceiling reached
24:                     result ← 0, break
25:                 qualifier, temp ← sum + product   → re-execute addition
26:                 if qualifier = False then
27:                     if error ← error + factor > n then   → exit if error ceiling reached
28:                         result ← 0, break
29:             if error > 0 then    → decrement error
30:                 error ← error – 1
31:             sum ← temp, temp ← 0
32:     result ← sum
33:     failure ← False
34: end procedure
```
*Algorithm 3: Example for a reliably executed convolution kernel.*

We implement this in Python and C, integrating the algorithm into the TensorFlow execution path on a standard Desktop [3]. We measure the following characteristics for execution of the first AlexNet convolution layer, that is calculation of 96 feature maps by 96 11*11*3 filters. For reference, native TensorFlow execution achieves this in 0.05s. As Algorithm 2 performs two multiplications and a comparison, the measured times in Python relative to each other are feasible. We expect these execution times to be substantially reduced when the operations are implementation in hardware. In comparison, a naïve version of the SAX algorithm to determine shape completes in 1.942 seconds on the same platform.

---
[3] Lenovo ThinkStation P330, Intel® i9-9900 CPU @ 3.10GHz
Ubuntu 23.04, Linux 6.2.0-39-generic

*Table 1: Execution time for the reliable convolution algorithm (Algorithm 3) with and without reliably executed operations.*

|  | Multiplication (Algorithm 1) | Redundant Multiplication (Algorithm 2) |
|---|---|---|
| Execution in Python | 301.91 s | 648.87 s |

In comparison, a naïve version of the SAX algorithm to determine shape completes in 1.942 seconds on the same platform. We find, unsurprisingly, that the implementation of reliable execution features in a neural network is feasible but requires some optimization. We also find that the principle of a hybrid neural network is feasible.

## V. CONCLUSION

### A. Results

We have proposed and reasoned a hybrid CNN for model-dependent reliable execution, proposed an architecture and produced first results. The advantage of our proposal is that we can reduce the necessary reliable execution to limits that a dependable model determines rather than just reliably executing an entire CNN or maintaining two parallel yet independent execution paths. We conserve both footprint and computational power. We have limited ourselves to one convolution layer in this body of work because our reliable model only requires edge detection before going into highly specific processing as opposed to the generalization that a CNN normally performs. We believe it is worthwhile investigating under what conditions subsequent layers of the CNN can be harnessed to determine input compliance to a dependable model.

Another contribution we provide is the integration of a rollback in the case of at least a single error. We can subsequently adjust the number of errors required to report an error condition serious enough to consider the application irrecoverable. We see further work in two distinct directions.

### B. Future Work

CNN model creation is an iterative process that allows the developer to optimize the network to some key performance indicator. Introducing reliability constraints results in the requirement for a reliability and/or safety-oriented workflow. In the safety and dependability domain, this in turn demands that tools, that is very specific tool versions, become certified. The certification of these tools incurs a considerable expense and imposes inflexibility on tool maintenance and further development. In a fast-moving domain such as AI, a lightweight workflow supported by tools of restricted scope will be preferrable to rich tooling with associated costs and certification. We believe that focus should be placed on researching extensions to the ONNX standard to facilitate the platform-agnostic description of hybrid-CNNs.

Given the multitude of AI-acceleration platforms that have no built-in reliability features, we believe that experimentation with this and similar algorithms on these platforms is warranted. Our focus is integrating the algorithm with the FPGA-IP simulation framework so that we can experiment with different HW arithmetic logic units.